\documentclass[11pt,a4paper]{article}
\usepackage[hyperref]{ranlp2023}
\usepackage{times}
\usepackage{latexsym}

\usepackage{microtype}

\aclfinalcopy % Uncomment this line for the final submission
\usepackage{CJKutf8}
\usepackage{amssymb}
\usepackage{adjustbox}

\usepackage{makecell}

\title{Named Entity Recognition in Context:\\
Edit\_Dunhuang team Technical Report for EvaHan2025 NER Competition}

\author{
 \textbf{Colin Brisson\textsuperscript{1, 2}},
 \textbf{Ayoub Kahfy\textsuperscript{2}},
 \textbf{Marc Bui\textsuperscript{1}},
 \textbf{Frédéric Constant\textsuperscript{3}},
\\
 \textsuperscript{1}EPHE-Université PSL, Paris France,
 \textsuperscript{2}Badiane, Neufchâtel-Hardelot, France,
\\
 \textsuperscript{3}Université Côte d'Azur, Nice, France
\\
 \small{
   \textbf{Correspondence:} \href{mailto:colin.brisson@ephe.psl.eu}{colin.brisson@ephe.psl.eu}
 }
}

\begin{document}
\maketitle
\begin{abstract}
We present the Named Entity Recognition system developed by the Edit\_Dunhuang team for the EvaHan2025 competition. Our approach integrates three core components: (1) \texttt{Pindola}, a modern transformer-based bidirectional encoder pretrained on a large corpus of Classical Chinese texts; (2) a retrieval module that fetches relevant external context for each target sequence; and (3) a generative reasoning step that summarizes retrieved context in Classical Chinese for more robust entity disambiguation. Using this approach, we achieve an average F1 score of 85.58, improving upon the competition baseline by nearly 5 points.

\end{abstract}

\section{Introduction}

The EvaHan2025 competition aimed to evaluate the state-of-the-art in Named Entity Recognition (NER) for Classical Chinese texts. The evaluation was conducted using three distinct datasets, each containing different types of entities. Dataset A comprised texts from the Shiji, the historical records composed by Sima Qian during the late 2nd and early 1st centuries BCE. This dataset included annotations for six entity types: person names, geographical locations, book titles, official titles, country names, and temporal expressions. Dataset B featured more diverse excerpts drawn from the Twenty-Four Histories, the official dynastic histories of China, but contained annotations for only three entity types: person names, geographical locations, and temporal expressions. Dataset C differed significantly from the other datasets, consisting exclusively of medicinal texts annotated for six specialized entities: disease names, syndromes, medicinal formulas, decoction pieces, symptoms and acupuncture points.

Our participation in the competition is part of the Read\_Chinese (BnF-Datalab) and Edit\_Dunhuang (Biblissima+) projects, which aim to produce digital facsimile of the Chinese documents in the Pelliot collection of the Bibliothèque nationale de France and the Stein collection of the British Library. Both collections consist of documents—primarily manuscripts on paper—discovered in the early 20th century in Dunhuang (Gansu province) in northwest China. These documents provide crucial insights into the history of medieval China as well as the transmission of ideas before the adoption of woodblock printing. As part of these projects, we intend not only to transcribe the text of the manuscripts but also to convert the OCR output into a structured and richly annotated text.

Our strategy was built on three core ideas. First, we developed \texttt{Pindola}, a modern transformer-based bidirectional encoder pretrained on a large corpus of Classical Chinese texts. \texttt{Pindola} incorporates key enhancements that significantly improve the quality of learned representations. Second, recognizing that target sentences are not isolated fragments but parts of a broader, interconnected context, we enriched our input sequences with external contextual information to support more accurate annotations. Finally, we employed a reasoning generative model to refine and summarize this context, thereby improving the model’s capacity for precise entity recognition.

\section{Related Work}

Integrating external context has emerged as a powerful approach to mitigate hallucinations and enhance factual accuracy in generative large language models (LLMs). A prominent example is Retrieval-Augmented Generation (RAG), which supplies models with relevant external context as raw text. RAG has achieved state-of-the-art results on various open-domain question-answering benchmarks, outperforming both standalone generative models and specialized retrieval-and-reading pipelines \citep{RAG_for_knowledge_intensive_nlp_tasks}. Another promising strategy leverages structured data, notably knowledge graphs. This approach has been effective in improving entity recognition accuracy within traditional Chinese texts \citep{duan2025research}. However, generative LLMs still face significant challenges in Named Entity Recognition (NER), particularly in specialized domains such as historical texts \citep{de-toni-etal-2022-entities}, and show a tendency to distort input sequences \citep{Li2024EvaHan}.

Traditional bidirectional encoder-based NER systems typically analyze sentences independently, often overlooking their broader contextual relationships. Recent studies, however, have demonstrated that integrating relevant external context can substantially improve the performance of these models. For instance, \citet{Wang2021ExternalNER} showed that incorporating context led to an improvement exceeding 2 points over the same model without context on the WNUT-17 dataset \citep{derczynski-etal-2017-results}, a benchmark designed specifically for recognizing unusual or emerging entities.

Several transformer-based bidirectional encoders have been developed for Classical Chinese, notably the GujiBERT family \citep{wang2023gujibertgujigptconstructionintelligent}, whose use was mandatory in the competition's closed modality and served to establish the competition baseline. However, due to computational constraints, these models were adapted from architectures originally trained for modern Chinese. Consequently, their architectures and performance levels are limited by design decisions made nearly a decade ago. Recent studies indicate that targeted architectural refinements significantly improve learned representations \citep{Warner2024Smarter}, especially in low-resource scenarios \citep{Samuel2023BNCBERT}. Moreover, novel optimization methods, such as FlashAttention \citep{Dao2022FlashAttention}, have reduced the cost of training new language models from scratch.

\section{System}

\subsection{Model}

Our model, named \texttt{Pindola} after a disciple of the Buddha who was once admonished for misusing his powers to impress simple people, is a transformer-based bidirectional model. \texttt{Pindola} incorporates several state-of-the-art innovations: it uses FlashAttention v2 \citep{Dao2022FlashAttention} for efficient attention computation, a SentencePiece tokenizer \citep{kudo-richardson-2018-sentencepiece} with a vocabulary of 65,536 tokens, SwiGLU activation \citep{shazeer2020gluvariantsimprovetransformer} and AliBias positional encoding \citep{press2022trainshort} to handle long input sequences of up to 2048 tokens\footnote{A comprehensive description of the model architecture and training methodology will be provided in an upcoming publication.}. Two variants were developed:
\begin{itemize}
    \item \texttt{Pindola\_small}: 12 layers with approximately 135 million parameters.
    \item \texttt{Pindola\_large}: 28 layers with approximately 360 million parameters.
\end{itemize}

For the competition, we fine-tuned two specialized variants derived from \texttt{Pindola}:
\begin{itemize}
\item \texttt{Pindola\_retrieval}: This variant of \texttt{Pindola\_small} was independently fine-tuned using contrastive self-supervised learning to embed both the target and contextual sentences.
\item \texttt{Pindola\_NER}: Built upon \texttt{Pindola\_large}, this model is equiped with a token classification head featuring two layers of bidirectional long short-term memory (Bi-LSTM) followed by a conditional random field (CRF) layer.
\end{itemize}

\subsection{NER Data}

\begin{table}[h!]
\centering
\small
\begin{tabular}{lccc}
\hline
 & Dataset A & Dataset B & Dataset C \\
\hline
Train & 324 & 3,130 & 272 \\
Test  & 37  & 791   & 67  \\
\hline
Total & 361 & 3,921 & 339 \\
\hline
\end{tabular}
\caption{Competition datasets segmented into sequences $\leq$ 510 tokens.}
\label{tab:competition}
\end{table}

As shown in Table~\ref{tab:competition}, segmenting the datasets into sequences of 510 tokens or fewer reveals that their overall volume is relatively limited. Notably, Dataset B contains a higher number of sequences due to its inherently shorter segments. This limited data volume is generally considered insufficient for fine-tuning a deep model like \texttt{Pindola\_large} \citep{Mao2022UniPELT}.

To address this limitation, we compiled an additional pretraining dataset by aggregating various publicly available online resources. We standardized the annotations in this dataset using the scheme adopted for competition Dataset B. By merging these external resources with the competition datasets, we created a combined dataset of 12,007 annotated sequences. Although the original sources employed different annotation guidelines—resulting in a somewhat heterogeneous dataset—we plan to further refine and publicly release this resource to support future research on Classical Chinese NER.

\subsection{Contextual Data}

To generate contextual information, we leveraged the extensive corpus used to pretrain \texttt{Pindola}. This corpus consists of approximately 3 billion characters of carefully curated Classical Chinese texts. The documents were split into chunks of 510 tokens.
\subsection{System pipeline}

Our system is organized into three sequential stages.

\paragraph{Step 1: Context Retrieval.}  
First, we encode all available contextual sequences into vectors of dimension \(d = 768\) using \texttt{Pindola\_retrieval} and store them in a vector database. For a given target sentence \(T\), we compute its embedding \(\mathbf{t} \in \mathbb{R}^d\). We then perform a vector search using the L2 (Euclidean) distance,
\[
d(\mathbf{t}, \mathbf{c}) = \|\mathbf{t} - \mathbf{c}\|_2 = \sqrt{\sum_{i=1}^{d} (t_i - c_i)^2},
\]
to retrieve the top \(k = 20\) contextual sequences \(\{C_1, C_2, \dots, C_{20}\}\) that are most similar yet non-identical to \(T\).

\paragraph{Step 2: Context Summarization.}  
Next, a reasoning model is employed to generate concise summaries of the retrieved contexts. To avoid overfitting, for each target sentence in the training and validation sets, we derive a set of summaries \(\{S_1, S_2, \dots, S_5\}\); for the test set, only a single summary \(S_1\) is generated. We employ OpenAI's \texttt{o3-mini-2025-01-31} model via its API, which returns JSON-formatted outputs (see Appendix A for an example prompt and Appendix B for sample outputs).

\paragraph{Step 3: Token Classification.}  
Finally, \texttt{Pindola\_NER} performs token-level classification. The target sentence \(T\) is concatenated with one of its summarized contexts \(S\) using designated separation tokens, forming the composite input:
\[
X = \texttt{[CLS]} \, T \oplus \texttt{[SEP]} \oplus S \oplus \texttt{[SEP]}
\]
Although the entire sequence \(X\) is encoded jointly, only the token representations corresponding to \(T\) are used for classification. For each token \(x_i\) in \(T\), the predicted class is given by
\[
y_i = \arg\max_{c \in \mathcal{C}} f(x_i; \theta),
\]
where \(f(\cdot; \theta)\) is the token classification head of \texttt{Pindola\_NER} that maps the token's representation to a score over the entity classes, and \(\mathcal{C}\) is the set of entity classes.

\subsection{Training}
\begin{table}[ht]
\centering
\small
\begin{tabular}{lcc}
\hline
 & Pretraining & Fine-tuning \\
\hline
Input Sequence Length & 2048 & 2048 \\
Batch Size            & 32   & 8 \\
Optimizer             & AdamW & AdamW \\
$\epsilon$            & 1e-6  & 1e-6 \\
Encoder LR            & 1e-5  & 1e-5 \\
Head LR               & 1e-3  & 1e-3 \\
Encoder Weight Decay  & 1e-2  & 1e-2 \\
Head Weight Decay     & 1e-2  & 1e-2 \\
Dropout               & 0.2   & 0.3 \\
Warmup Steps          & 500   & 200 \\
\hline
\end{tabular}
\caption{Training parameters for \texttt{Pindola\_NER}.}
\label{tab:training_params}
\end{table}

Training of \texttt{Pindola\_NER} was conducted in two phases: an initial pretraining phase followed by fine-tuning on each competition dataset. Table~\ref{tab:training_params} summarizes the training parameters used in both phases.

\section{Results}

\begin{table*}[ht]
\centering
\resizebox{\textwidth}{!}{%
\begin{tabular}{l ccc ccc ccc ccc}
\hline
 & \multicolumn{3}{c}{Dataset A} & \multicolumn{3}{c}{Dataset B} 
 & \multicolumn{3}{c}{Dataset C} & \multicolumn{3}{c}{Overall} \\
 & Prec. & Rec. & F1 
 & Prec. & Rec. & F1 
 & Prec. & Rec. & F1 
 & Prec. & Rec. & F1 \\
\hline
\textbf{Initial Submission} 
& 82.16 & 78.51 & 80.29 
& 61.29 & 71.80 & 66.13 
& 46.85 & 59.18 & 52.30
& 60.90 & 69.45 & 64.90 \\
\textbf{Revised Submission} 
& \underline{87.44} & \underline{81.51} & \underline{84.37} 
& \underline{88.23} & \underline{89.34} & \underline{88.78} 
& \underline{78.66} & \underline{88.32} & \underline{83.21}
& \underline{84.47} & \underline{86.73} & \underline{85.58} \\
\textbf{Baseline} 
& 85.90 & 77.50 & 81.48
& 87.09 & 87.92 & 87.50
& 71.84 & 72.95 & 72.40
& 81.41 & 79.82 & 80.61 \\
\hline
\end{tabular}%
}
\caption{NER results on EvaHan2025 datasets (A, B, C) and Overall as evaluated by the competition organizers. Best results for each column are underlined.}
\label{tab:results}
\end{table*}

Table~\ref{tab:results} summarizes our system's performance on the EvaHan2025 datasets as evaluated by the competition organizers. An issue during data preparation led to suboptimal performance in our initial submission (Initial Submission). After the competition, we submitted a revised version (Revised Submission) that incorporated the necessary fixes, resulting in significant improvements. Specifically, the overall average F1 score increased to 85.58, nearly 5 points above the baseline.

\section{Ablation study}

\begin{table}[t!]
\centering
\begin{adjustbox}{max width=\linewidth}
\footnotesize
\begin{tabular}{lccc}
\hline
 & Dataset A & Dataset B & Dataset C \\
\hline
\textbf{w/ Context}      & \underline{$83.29 \pm 1.07$}  & \underline{$89.10 \pm 0.45$}  & \underline{$82.89 \pm 1.45$}  \\
\textbf{w/o Context}      & $83.68 \pm 0.72$  & $88.69 \pm 0.18$  & $83.09 \pm 0.79$  \\
\textbf{w/o Pretraining}  & $83.60 \pm 0.50$  & $88.65 \pm 0.48$  & $82.29 \pm 0.66$  \\
\hline
\end{tabular}
\end{adjustbox}
\caption{Average F1 Scores computed over three runs (Seeds 42, 123, and 2025). For each dataset, the experiment that achieved the best run is underlined.}
\label{tab:f1_results}
\end{table}

To assess the contributions of external context and the pretraining phase, we evaluated the model under three configurations: (1) with external context during both the pretraining phase and competition dataset training (as in our Revised Submission, denoted as "w/ Context" in Table~\ref{tab:f1_results}), (2) without external context in either phase (denoted as "w/o Context"), and (3) with external context applied only during competition dataset training, thereby omitting the pretraining phase (denoted as "w/o Pretraining"). For each configuration, we conducted experiments using three different random seeds (42, 123, and 2025). The results are summarized in Table~\ref{tab:f1_results}.

Our evaluation shows that incorporating external context generally improves performance across all three datasets, though it also increases variability—evidenced by standard deviations exceeding 1 on Datasets A and C. Notably, only Dataset B exhibits a consistent average improvement when context is added, which may be due to a closer alignment between our pretraining dataset and Dataset B. Furthermore, models trained without the pretraining phase tend to perform worst, albeit with only a modest decline. Overall, while these differences suggest trends in how each component affects performance, the high variability warrants cautious interpretation.

\section{Analysis}

The ablation study suggests that, in the current configuration, both external context integration and pretraining yield only minimal improvements. This may be because the entities across these three datasets exhibit little ambiguity—allowing the model to distinguish them effectively based solely on the linguistic context of the sentence—or because the generated context summaries do not provide sufficient additional information. In any case, these findings imply that similar performance could potentially be achieved without the need for external context or supplementary pretraining data.

As anticipated, our model achieves the highest overall performance on Dataset B, which features the simplest labeling scheme and closely aligns with the pretraining dataset. The lowest performance is observed on Dataset C, likely due to the medical texts being underrepresented in the \texttt{Pindola} pretraining dataset. Nonetheless, Dataset C also exhibits the largest improvement over the baseline, which underscores that \texttt{Pindola} constitutes a significant advancement over existing models.

\section{Conclusion}

In this work, we introduced our NER system developed for the EvaHan2025 competition, which achieved an overall average F1 score of 85.58, significantly surpassing the competition baseline. This performance highlights the advancements brought by \texttt{Pindola}, our modern transformer-based bidirectional encoder designed specifically for Classical Chinese. Interestingly, our experiments suggest that comparable results may be attainable without relying on external context or extensive pretraining on large corpora, thereby simplifying future applications. These findings open promising avenues for further research into more efficient yet effective approaches to NER in low-resource and historical language settings.

\section{Limitations}

Due to the constraints of the competition, we were unable to fully optimize every component of our system or conduct an exhaustive search for the best hyperparameters. Consequently, further optimization could potentially yield improved performance. Moreover, the modest benefits observed from incorporating external context may be attributed to limitations in our retrieval and summarization modules. Future work should explore alternative retrieval strategies and experiment with varying approaches to context integration—such as using minimal or even no summarization—to better understand and enhance the impact of external context.

\section*{Acknowledgments}

The authors express their gratitude to the Bibliothèque nationale de France and the British Library for their support. This work was supported by the French National Research Agency under the Investissements d’avenir initiative (reference ANR-21-ESRE-0005, ÉquipEx Biblissima+). This work was granted access to the HPC resources of IDRIS under the allocation 2025-103542 made by GENCI.

\bibliographystyle{acl_natbib}
\bibliography{references}

\begin{thebibliography}{14}
\providecommand{\natexlab}[1]{#1}

\bibitem[{Dao et~al.(2022)Dao, Fu, Ermon, Rudra, and R{\'e}}]{Dao2022FlashAttention}
Tri Dao, Daniel~Y. Fu, Stefano Ermon, Atri Rudra, and Christopher R{\'e}. 2022.
\newblock Flash{A}ttention: Fast and memory-efficient exact attention with {IO}-awareness.
\newblock In \emph{Advances in Neural Information Processing Systems}, volume~35, pages 16344--16359. Curran Associates, Inc.

\bibitem[{De~Toni et~al.(2022)De~Toni, Akiki, De~La~Rosa, Fourrier, Manjavacas, Schweter, and Van~Strien}]{de-toni-etal-2022-entities}
Francesco De~Toni, Christopher Akiki, Javier De~La~Rosa, Cl{\'e}mentine Fourrier, Enrique Manjavacas, Stefan Schweter, and Daniel Van~Strien. 2022.
\newblock \href {https://doi.org/10.18653/v1/2022.bigscience-1.7} {Entities, dates, and languages: Zero-shot on historical texts with t0}.
\newblock In \emph{Proceedings of BigScience Episode {\#}5 -- Workshop on Challenges {\&} Perspectives in Creating Large Language Models}, pages 75--83, virtual+Dublin. Association for Computational Linguistics.

\bibitem[{Derczynski et~al.(2017)Derczynski, Nichols, van Erp, and Limsopatham}]{derczynski-etal-2017-results}
Leon Derczynski, Eric Nichols, Marieke van Erp, and Nut Limsopatham. 2017.
\newblock \href {https://doi.org/10.18653/v1/W17-4418} {Results of the {WNUT}2017 shared task on novel and emerging entity recognition}.
\newblock In \emph{Proceedings of the 3rd Workshop on Noisy User-generated Text}, pages 140--147, Copenhagen, Denmark. Association for Computational Linguistics.

\bibitem[{Duan et~al.(2025)Duan, Zhou, Li, Qin, Wang, Kan, and Hu}]{duan2025research}
Yuchen Duan, Qingqing Zhou, Yu~Li, Chi Qin, Ziyang Wang, Hongxing Kan, and Jili Hu. 2025.
\newblock \href {https://doi.org/10.3389/fmed.2024.1512329} {Research on a traditional chinese medicine case-based question-answering system integrating large language models and knowledge graphs}.
\newblock \emph{Frontiers in Medicine}, 11:1512329.
\newblock ECollection 2024.

\bibitem[{Kudo and Richardson(2018)}]{kudo-richardson-2018-sentencepiece}
Taku Kudo and John Richardson. 2018.
\newblock \href {https://doi.org/10.18653/v1/D18-2012} {{S}entence{P}iece: A simple and language independent subword tokenizer and detokenizer for neural text processing}.
\newblock In \emph{Proceedings of the 2018 Conference on Empirical Methods in Natural Language Processing: System Demonstrations}, pages 66--71, Brussels, Belgium. Association for Computational Linguistics.

\bibitem[{Lewis et~al.(2020)Lewis, Perez, Piktus, Petroni, Karpukhin, Goyal, K\"{u}ttler, Lewis, Yih, Rockt\"{a}schel, Riedel, and Kiela}]{RAG_for_knowledge_intensive_nlp_tasks}
Patrick Lewis, Ethan Perez, Aleksandra Piktus, Fabio Petroni, Vladimir Karpukhin, Naman Goyal, Heinrich K\"{u}ttler, Mike Lewis, Wen-tau Yih, Tim Rockt\"{a}schel, Sebastian Riedel, and Douwe Kiela. 2020.
\newblock Retrieval-augmented generation for knowledge-intensive nlp tasks.
\newblock In \emph{Proceedings of the 34th International Conference on Neural Information Processing Systems}, NIPS '20, Red Hook, NY, USA. Curran Associates Inc.

\bibitem[{Li et~al.(2024)Li, Chang, Xu, Feng, Xu, Qu, Shen, and Wang}]{Li2024EvaHan}
Bin Li, Bolin Chang, Zhixing Xu, Minxuan Feng, Chao Xu, Weiguang Qu, Si~Shen, and Dongbo Wang. 2024.
\newblock Overview of {E}va{H}an2024: The first international evaluation on ancient chinese sentence segmentation and punctuation.
\newblock In \emph{Proceedings of the Third Workshop on Language Technologies for Historical and Ancient Languages (LT4HALA) @ LREC-COLING 2024}, pages 229--236. ELRA and ICCL.

\bibitem[{Mao et~al.(2022)Mao, Mathias, Hou, Almahairi, Ma, Han, Yih, and Khabsa}]{Mao2022UniPELT}
Yuning Mao, Lambert Mathias, Rui Hou, Amjad Almahairi, Hao Ma, Jiawei Han, Wen-tau Yih, and Madian Khabsa. 2022.
\newblock {UniPELT}: A unified framework for parameter-efficient language model tuning.
\newblock In \emph{Proceedings of the 60th Annual Meeting of the Association for Computational Linguistics (Volume 1: Long Papers)}, pages 6253--6264. Association for Computational Linguistics.

\bibitem[{Press et~al.(2022)Press, Smith, and Lewis}]{press2022trainshort}
Ofir Press, Noah~A. Smith, and Mike Lewis. 2022.
\newblock Train short, test long: Attention with linear biases enables input length extrapolation.
\newblock In \emph{International Conference on Learning Representations (ICLR 2022)}.

\bibitem[{Samuel et~al.(2023)Samuel, Kutuzov, {\O}vrelid, and Velldal}]{Samuel2023BNCBERT}
David Samuel, Andrey Kutuzov, Lilja {\O}vrelid, and Erik Velldal. 2023.
\newblock Trained on 100 million words and still in shape: {BERT} meets british national corpus.
\newblock In \emph{Findings of the Association for Computational Linguistics: EACL 2023}, pages 1954--1974. Association for Computational Linguistics.

\bibitem[{Shazeer(2020)}]{shazeer2020gluvariantsimprovetransformer}
Noam Shazeer. 2020.
\newblock \href {https://arxiv.org/abs/2002.05202} {Glu variants improve transformer}.
\newblock \emph{Preprint}, arXiv:2002.05202.

\bibitem[{Wang et~al.(2023)Wang, Liu, Zhao, Shen, Liu, Li, Hu, Wu, Lin, Zhao, and Wang}]{wang2023gujibertgujigptconstructionintelligent}
Dongbo Wang, Chang Liu, Zhixiao Zhao, Si~Shen, Liu Liu, Bin Li, Haotian Hu, Mengcheng Wu, Litao Lin, Xue Zhao, and Xiyu Wang. 2023.
\newblock \href {https://arxiv.org/abs/2307.05354} {Gujibert and gujigpt: Construction of intelligent information processing foundation language models for ancient texts}.
\newblock \emph{Preprint}, arXiv:2307.05354.

\bibitem[{Wang et~al.(2021)Wang, Jiang, Bach, Wang, Huang, Huang, and Tu}]{Wang2021ExternalNER}
Xinyu Wang, Yong Jiang, Nguyen Bach, Tao Wang, Zhongqiang Huang, Fei Huang, and Kewei Tu. 2021.
\newblock Improving named entity recognition by external context retrieving and cooperative learning.
\newblock In \emph{Proceedings of the 59th Annual Meeting of the Association for Computational Linguistics and the 11th International Joint Conference on Natural Language Processing (Volume 1: Long Papers)}, pages 1800--1812. Association for Computational Linguistics.

\bibitem[{Warner et~al.(2024)Warner, Chaffin, Clavi{\'e}, Weller, Hallstr{\"o}m, Taghadouini, Gallagher, Biswas, Ladhak, Aarsen, Cooper, Adams, Howard, and Poli}]{Warner2024Smarter}
Benjamin Warner, Antoine Chaffin, Benjamin Clavi{\'e}, Orion Weller, Oskar Hallstr{\"o}m, Said Taghadouini, Alexis Gallagher, Raja Biswas, Faisal Ladhak, Tom Aarsen, Nathan Cooper, Griffin Adams, Jeremy Howard, and Iacopo Poli. 2024.
\newblock Smarter, better, faster, longer: A modern bidirectional encoder for fast, memory efficient, and long context finetuning and inference.
\newblock \emph{arXiv preprint arXiv:2412.13663}.

\end{thebibliography}

\newpage

\appendix
\section{Sample Prompt for Context Summarization (Dataset A)}
\label{sec:appendix_prompt}

\noindent\textit{(Target sentence and generated context omitted for brevity)}\\[1ex]

\subsection*{Developer Instructions}
\begin{quote}
You are an expert in Chinese history and literature. You provide clear, concise answers in Classical Chinese and can leverage additional context to enhance your explanations.
\end{quote}

\subsection*{User Prompt}
\begin{quote}
Read carefully the following text and extract all clues that can help identify the following entities in the target sentence:
\begin{CJK*}{UTF8}{bsmi}
\begin{itemize}
    \item \textbf{Person name} (e.g., 荆軻, 伏羲)
    \item \textbf{Geographical location} (e.g., 長平, 黄河)
    \item \textbf{Book title} (e.g., 易, 易經)
    \item \textbf{Official title} (e.g., 中大夫)
    \item \textbf{Country name} (e.g., 秦)
    \item \textbf{Time expression} (e.g., 三月, 丙戌, 丁亥)
\end{itemize}
\end{CJK*}
\textbf{Target Sentence:} \ldots \\
\textbf{Context:} \ldots \\[1ex]
Provide three distinct explanations of your findings in Classical Chinese. Output your responses as a JSON array of objects, with each object containing a brief textual explanation.''
\end{quote}

\section{Target sentence and generated context sentence (Dataset A)}
\label{sec:appendix_context}

\begin{CJK*}{UTF8}{bsmi}
\textbf{Target sentence:} 及封中大謁者張釋爲建侯，吕榮爲祝侯。諸中宦者令丞皆爲關內侯，食邑五百戶。七月中，高后病甚，迺令趙王吕禄爲上將軍，軍北軍；吕王産居南軍。吕后誡產、禄曰：「高帝已定天下，與大臣約，曰『非劉氏王者，天下共擊之』。今吕氏王，大臣弗平。我卽崩，帝年少，大臣恐爲變。必據兵衞宮，慎毋送喪，毋爲人所制。」辛巳，高后崩，遺詔賜諸侯王各千金，將相列侯郎吏皆以秩賜金。大赦天下。以吕王産爲相國，以吕禄女爲帝后。高后已葬，以左相審其爲帝太傅。朱侯劉章有氣力，東侯興居其弟也。皆齊王弟，居長安。當是時，諸吕用事擅權，欲爲亂，畏高帝故大臣絳、灌等，未敢發。朱侯婦，吕禄女，陰知其謀。恐見誅，迺陰令人告其兄齊王，欲令發兵西，誅諸吕而立。朱侯欲從中與大臣爲應。齊王欲發兵，其相弗聽。八月丙午，齊王欲使人誅相，相召平迺反，舉兵欲圍王，王因殺其相，遂發兵東，詐奪琅王兵，并將之而西。語在齊王語中。齊王迺遺諸侯王書曰：「高帝平定天下，王諸子弟，悼王王魏。悼王薨，孝帝使留侯良立臣爲齊王。孝惠崩，高后用事，春秋高，聽諸吕，擅廢帝更立，又比殺三趙王，滅梁、趙、燕以王諸吕，分魏爲四。忠臣進諫，上惑亂弗聽。今高后崩，而帝春秋富，未能治天下，固恃大臣諸侯。

\textbf{Context sentence:} 夫文中所載，先有封侯之事。其曰「中大謁者張釋」，「張釋」乃人名；「吕榮」亦人也。又「諸中宦者令丞」乃官職稱謂。又時曰「七月中」，又記「辛巳」之日，高后既病且崩，均顯時辰。此外，「建侯」、「祝侯」皆封爵。故人名、官職、時令，皆各有所示。
\end{CJK*}

\end{document}